\documentclass[lettersize,journal]{IEEEtran}
\usepackage{amsmath,amsfonts}
\usepackage{algorithmic}
\usepackage{algorithm}
\usepackage{array}
\usepackage{textcomp}
\usepackage{stfloats}
\usepackage{url}
\usepackage{verbatim}
\usepackage{graphicx}
\usepackage{cite}
\usepackage{color}
\usepackage{booktabs}
\usepackage{multirow}
\usepackage{subfigure}
\usepackage{bbding}
\usepackage{threeparttable}
\usepackage{makecell}
\usepackage{diagbox}
\usepackage{adjustbox}
\usepackage{array}
\usepackage{verbatim}
\usepackage{dsfont}
\usepackage{pifont}

\newcommand{\hlb}[1]{{\color{blue}{#1}}}

\newcommand{\bX}{\mathbf{X}}
\newcommand{\bx}{\mathbf{x}}

\newcommand{\bZ}{\mathbf{Z}}
\newcommand{\bW}{\mathbf{W}}

\newcommand{\bL}{\mathbf{L}}
\newcommand{\bp}{\mathbf{p}}
\newcommand{\bP}{\mathbf{P}}

\begin{document}

\title{CurveFormer++: 3D Lane Detection by Curve Propagation with Temporal Curve Queries and Attention}

\author{Yifeng Bai$^{ 1,2\dagger \ddagger}$, Zhirong Chen$^{3 \dagger}$, Pengpeng Liang$^{4}$, Bo Song$^{1*}$ and Erkang Cheng$^{3 *}$ 
    \thanks{$^{1}$Institute of Intelligent Machines, HFIPS, Chinese Academy of Sciences, Hefei, 230031, China.}
    \thanks{$^{2}$University of Science and Technology of China, Hefei, 230026, China.}
	\thanks{$^{3}$NullMax, Shanghai, 201210, China.}
    \thanks{$^{4}$School of Computer and Artificial Intelligence, Zhengzhou University, 450001, China.}
    \thanks{$\dagger$ Equal contribution. $\ddagger$ Work done during an internship at NullMax.}
	\thanks{*Corresponding author. chengerkang@nullmax.ai.}}

\maketitle

\begin{abstract}

In autonomous driving, accurate 3D lane detection using monocular cameras is important for downstream tasks. Recent CNN and Transformer approaches usually apply a two-stage model design. The first stage transforms the image feature from a front image into a bird's-eye-view (BEV) representation. Subsequently, a sub-network processes the BEV feature to generate the 3D detection results. However, these approaches heavily rely on a challenging image feature transformation module from a perspective view to a BEV representation. In our work, we present CurveFormer++, a single-stage Transformer-based method that does not require the view transform module and directly infers 3D lane results from the perspective image features. Specifically, our approach models the 3D lane detection task as a curve propagation problem, where each lane is represented by a curve query with a dynamic and ordered anchor point set. By employing a Transformer decoder, the model can iteratively refine the 3D lane results. A curve cross-attention module is introduced to calculate similarities between image features and curve queries. To handle varying lane lengths, we employ context sampling and anchor point restriction techniques to compute more relevant image features. Furthermore, we apply a temporal fusion module that incorporates selected informative sparse curve queries and their corresponding anchor point sets to leverage historical information. In the experiments, we evaluate our approach on two publicly real-world datasets. The results demonstrate that our method provides outstanding performance compared with both CNN and Transformer based methods. We also conduct ablation studies to analyze the impact of each component.

\end{abstract}


\section{Introduction}

\IEEEPARstart{L}ane detection plays a crucial role in autonomous driving perception systems, providing accurate lane information from a frontal camera for static traffic scenes. By leveraging lane detection results in the ego vehicle coordinate system, various essential driving features can be developed, including basic ADAS features such as lane keeping assist (LKA), lane departure warning (LDW), and more advanced functionalities like intelligent cruise control (ICC) and navigation on pilot (NOP). These features contribute to enhancing driving assistance and automation capabilities, ultimately improving the safety and convenience of autonomous vehicles.

\begin{figure*}[!htb]
  \centering
    \subfigure[Image prediction \& Post-processing.]{\includegraphics[width=0.4\textwidth]{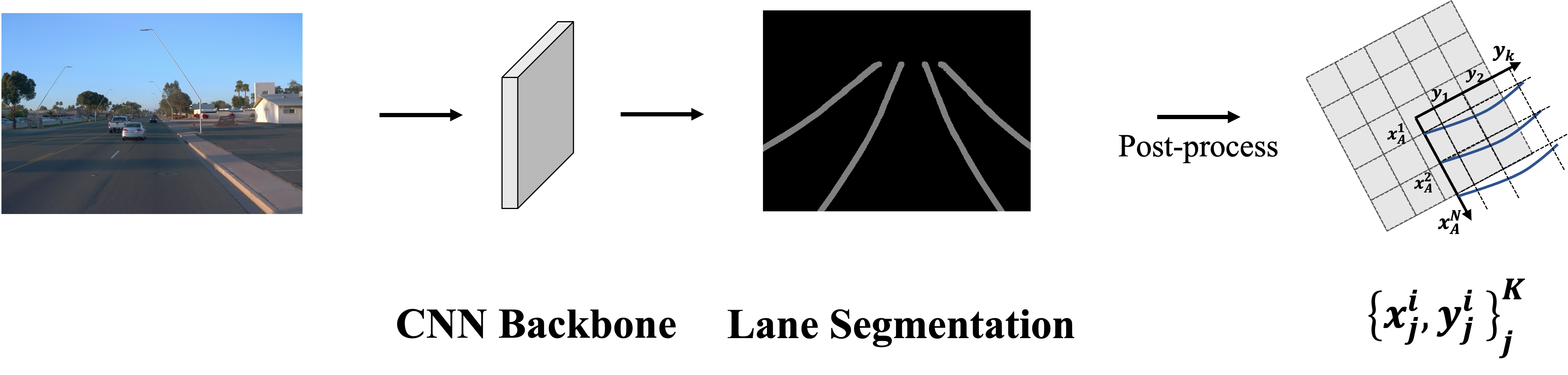}} \hspace{5mm} 
    \subfigure[CNN-based dense BEV \& Prediction.]{\includegraphics[width=0.4\textwidth]{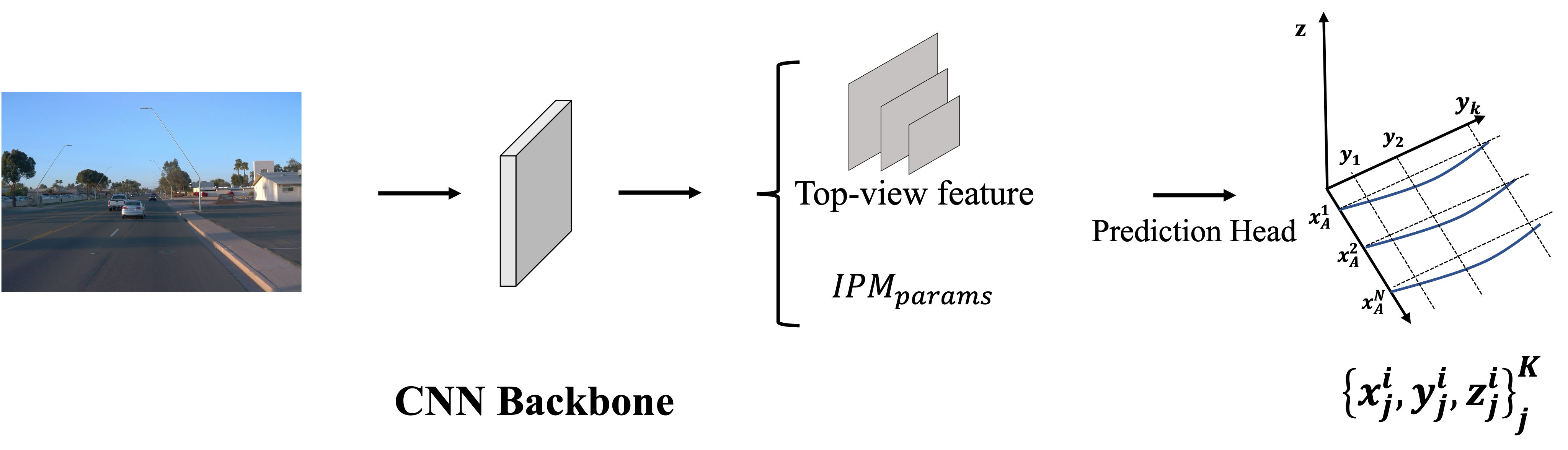}} \\
    \subfigure[Transformer-based dense BEV \& Prediction.]{\includegraphics[width=0.4\textwidth]{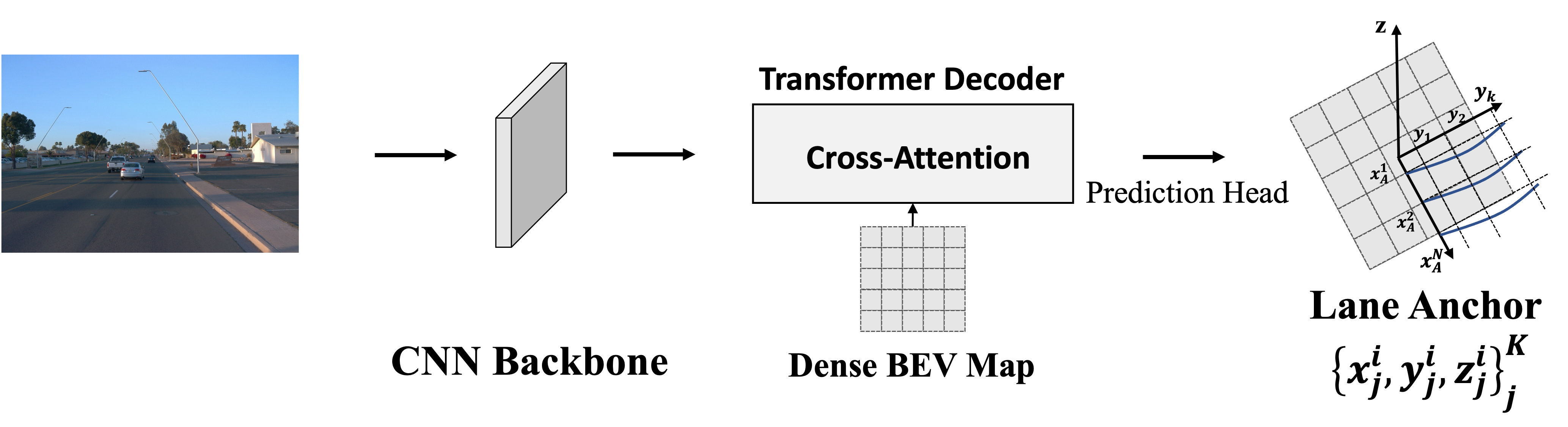}} \hspace{5mm} 
    \subfigure[Transformer-based sparse 3D lane detection by curve query.]{\includegraphics[width=0.4\textwidth]{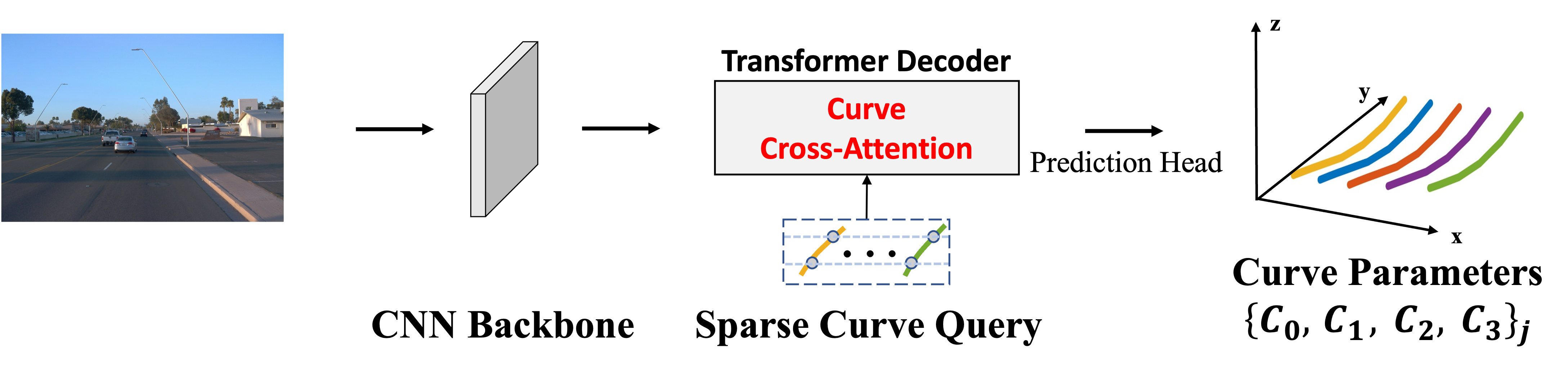}} \\
  \caption{Comparisons of different 3D lane detection pipelines. (a) 2D image prediction and post-processing; (b) 3D lane detection with camera extrinsic prediction; (c) Transformer-based dense BEV map construction and 3D lane prediction; (d) Our proposed CurveFormer++, directly provides 3D lane parameters by sparse curve queries with curve cross-attention mechanism in Transformer decoder.} 
    \label{fig:fig_compare_3Dmethods}
\end{figure*}

Early studies on lane detection primarily focused on the image space, treating it as a semantic segmentation task~\cite{pan2018spatial, neven2018towards, hou2019learning, zheng2021resa,10182290,9872124} or utilizing line regression techniques~\cite{chen2019pointlanenet, li2019line,ko2021key, qin2020ultra,xu2020curvelane, tabelini2021keep,wang2022keypoint, zheng2022clrnet,9768189,10196360}. 
For instance, the CNN-based segmentation method, SCNN~\cite{pan2018spatial} distinguishes lane lines from the traffic background from the input image, while the regression-based method Ultra-Fast~\cite{qin2020ultra} aims to identify key points on the lane lines on the image-space.
However, for downstream tasks such as planning and control, it is preferable to have lane representations in the form of curve parameters in 3D space. Therefore, a post-processing step is required to transfer 2D lane results from the image space to the ego-vehicle coordinate system, as illustrated in Fig.~\ref{fig:fig_compare_3Dmethods} (a).
Unfortunately, projecting the lanes from the image plane to the bird's-eye-view (BEV) perspective often leads to an error propagation problem due to the lack of depth information and accurate real-time camera extrinsic parameters. Moreover, typical post-processing steps such as clustering and curve fitting methods tend to be complex and time-consuming, making the lane detection approaches less robust and less suitable for realistic perception systems.


To address the limitations of post-processing in two-stage methods, CNN-based approaches have been proposed for end-to-end 3D lane detection tasks~\cite{garnett20193d,efrat20203d,guo2020gen,yan2022once, ai2023ws}. These methods employ Inverse Projective Mapping (IPM) to transfer image feature maps onto the ground plane. For example, 3D-LaneNet~\cite{garnett20193d}, shown in Fig.\ref{fig:fig_compare_3Dmethods} (b), uses an anchor-based 3D lane representation and predicts camera pose to project 2D features using IPM. Similarly, Gen-LaneNet\cite{guo2020gen} proposes a virtual top view that aligns IPM-projected BEV features with real-world lanes.
Another approach for CNN-based 3D lane detection involves depth estimation and the integration of this information. For instance, ONCE~\cite{yan2022once} performs 2D lane semantic segmentation and depth estimation, integrating these results to derive 3D lanes. 
However, the requirement of accurate camera pose estimation or depth estimation can potentially lead to the loss of lane height, thereby impacting the robustness of these methods, especially in scenarios where the flat ground assumption does not hold.

In recent years, Transformer-based methods have made significant advancements in various computer vision and robotic tasks, showcasing remarkable success~\cite{dosovitskiy2020image, carion2020end, wang2022detr3d, peng2023bevsegformer, pan2023slide,zhao2024ssir,huang2023understanding}. Originally introduced to the field of object detection by DETR~\cite{carion2020end}, Transformer-based approaches have gained popularity due to their ability to eliminate the need for post-processing steps, enabling direct inference of outputs from input images. 
Similarly, Transformer-based 3D lane detection methods adopt similar principles to calculate 3D lane results from input images.
These methods first construct a dense Bird's Eye View (BEV) map through view transformation and then compute the 3D lane results from the intermediate BEV feature map using cross-attention in decoder layers. For instance, PersFormer~\cite{chen2022persformer} constructs a dense BEV query and utilizes Transformers to interact queries from the BEV with image features (as shown in Fig.~\ref{fig:fig_compare_3Dmethods} (c)). However, despite their efforts to leverage Transformers for 3D lane detection, the absence of image depth or BEV map height limits their performance by hindering the acquisition of features that precisely correspond to the query. Consequently, these approaches have the potential to constrain the representation of vertical information, especially in 3D lane detection scenarios where the close range corresponds to a flat road, while the far distance may involve uphill or downhill sections.

\begin{figure}[t!]
	\centering
	\includegraphics[width=0.6\linewidth]{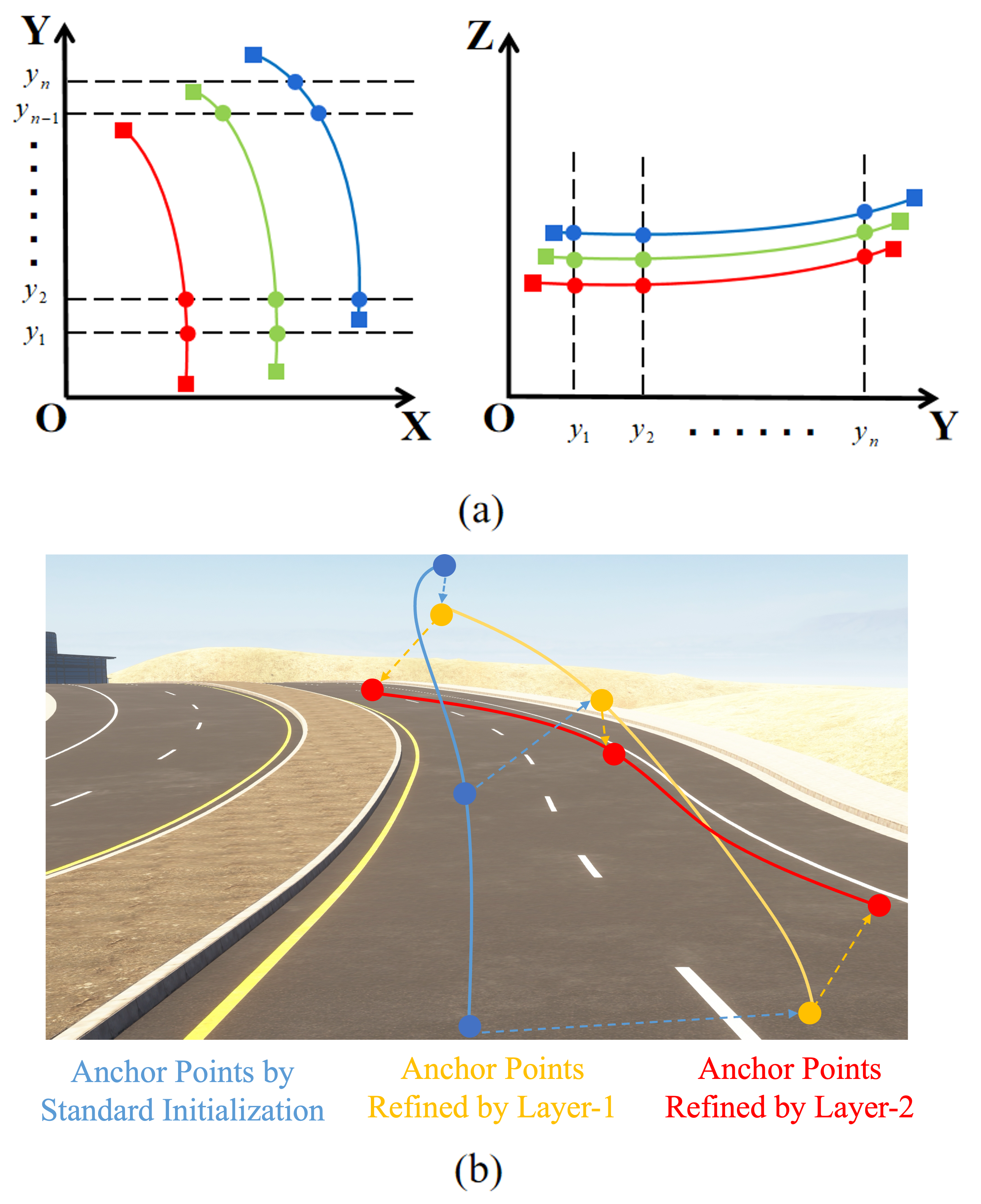}
	\caption{Illustration of the curve query representation with dynamic anchor point set in the X-O-Y plane and Z-O-Y plane. (a)  the iterative refinement visualization in the image view (b). Each dynamic anchor point set initially follows a standard normal distribution.}
	\label{fig:representation}
\end{figure}

In order to tackle the aforementioned challenges, we propose CurveFormer++, an enhanced
Transformer-based approach for 3D lane detection (Fig.\ref{fig:fig_compare_3Dmethods} (d)). In our method, lanes are represented as sparse curve queries and their corresponding lane confidence, two polynomials, and start and end points (Fig.\ref{fig:representation} (a)). Taking inspiration from DAB-DETR~\cite{liu2022dab}, we introduce a set of 3D dynamic anchor points to facilitate the interaction between curve queries and image features. By incorporating height information by using camera extrinsic parameters, the 3D anchor points enable us to accurately align the image features corresponding to each anchor point. Also, the dynamic anchor point set is iteratively refined within the sequence of Transformer decoders.
In addition, we introduce a novel curve cross-attention module in the decoder layer to investigate the effect of curve queries and dynamic anchor point sets. Different from standard Deformable-DETR~\cite{zhu2020deformable } that directly predicts sampling offsets from the query, we introduce a context sampling unit to predict offsets from the combination of reference features and queries to guide sampling offsets learning. 
To enable the extraction of more accurate features for lanes with varying lengths, we employ a dynamic anchor point range prediction as a restriction during feature sampling steps. 
In addition, an auxiliary segmentation branch is adopted to enhance the shared CNN backbone. In this way, our design of CurveFormer lends itself to 3D lane detection.

\begin{figure}[t!]
 \centering
  \begin{tabular}{c}
   \includegraphics[width=0.85\linewidth]{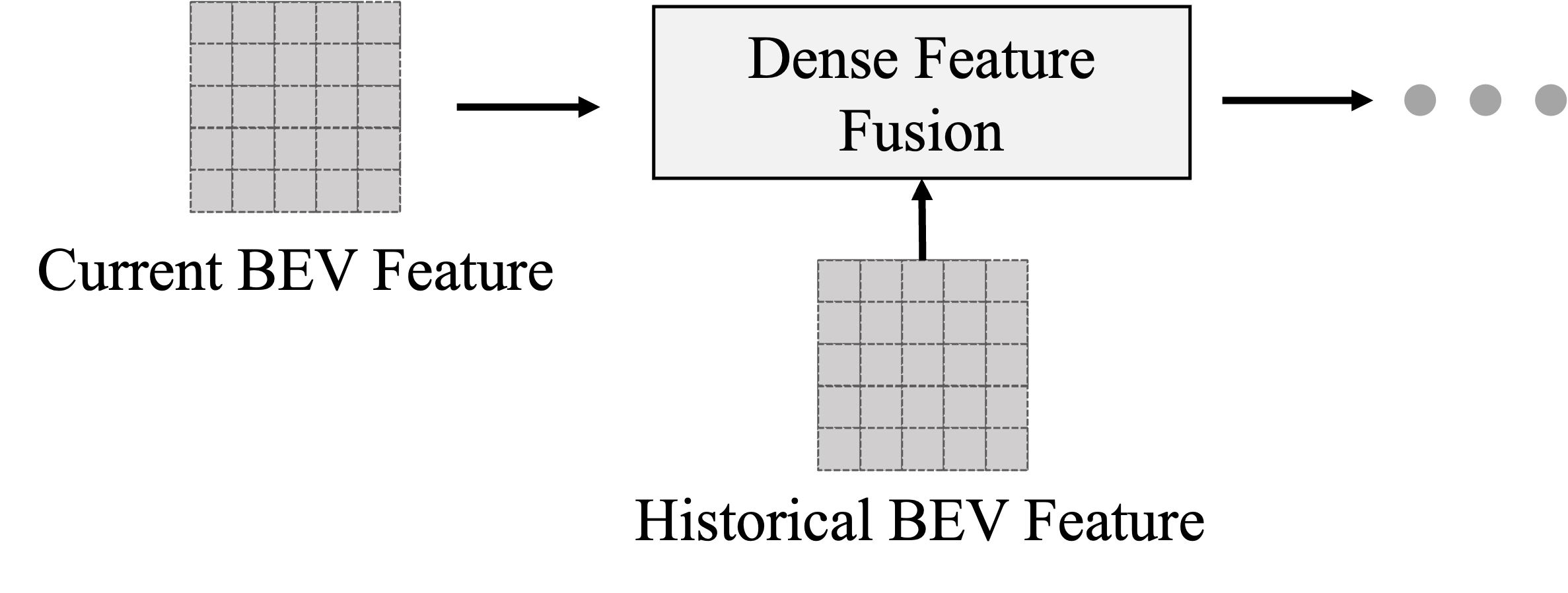}\\
   \begin{scriptsize}  
   (a) Fuse of temporal dense BEV map. 
   \end{scriptsize} 
   \\
   \includegraphics[width=0.85\linewidth]{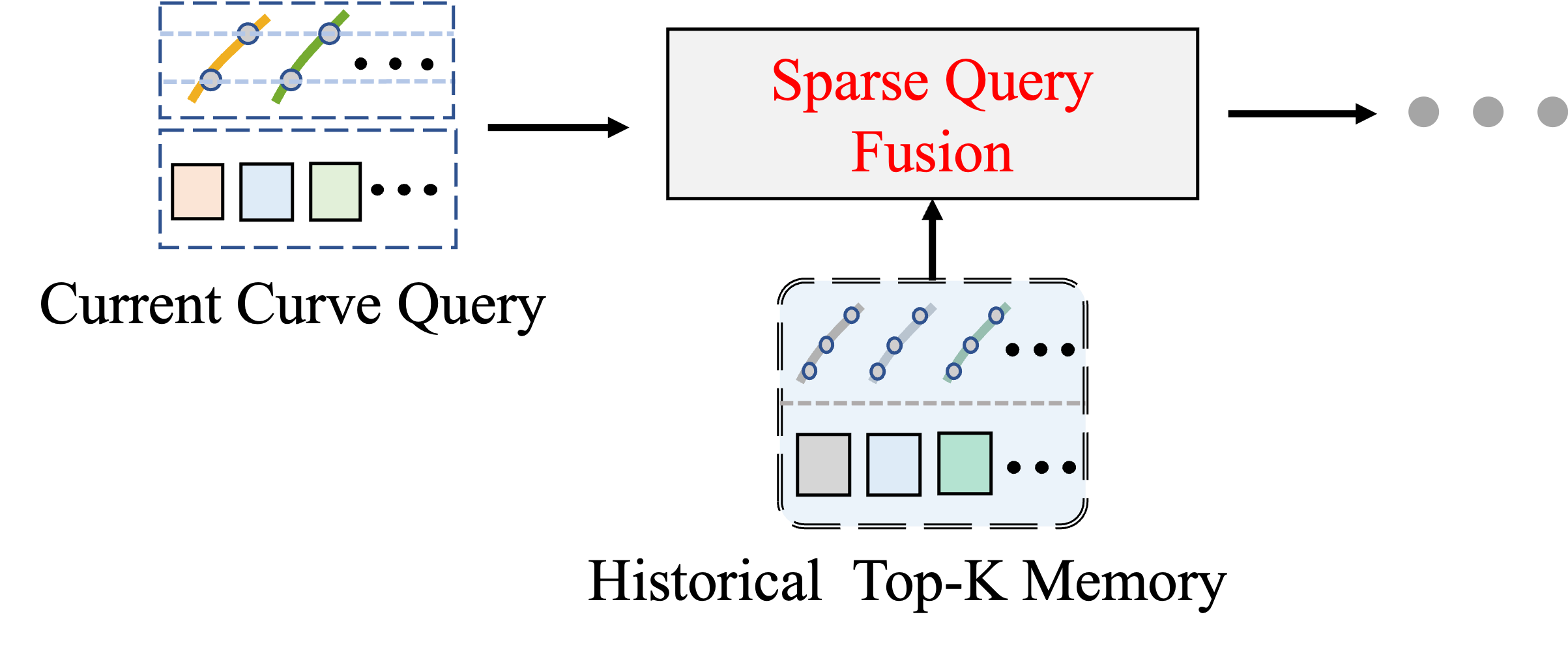}\\
   \begin{scriptsize}  
   (b) Fuse of temporal sparse curve query and anchor points.
   \end{scriptsize}  \\
 \end{tabular}
 \caption{Comparisons of different Transformer-based temporal information fusion approaches for 3D lane detection.}
 \label{fig:fig_compare_4Dmethods}
\end{figure}

Temporal information from historical frames plays a crucial role in advancing 3D perception in autonomous driving. Recently, Transformer-based BEV methods~\cite{li2022bevformer,liang2022bevfusion} fuse BEV feature maps built from multi-frame image features into a unified BEV space to provide temporal information. These methods achieve significant performance in object detection and static traffic scene understanding tasks compared with single-frame methods. 
For the 3D lane detection task, STLane3D~\cite{wang2022spatio} applies a similar idea to fuse dense BEV features from previous observation, as illustrated in Fig.~\ref{fig:fig_compare_4Dmethods} (a).
Despite the advantages of BEV feature fusion, the dense feature sampling necessary for the image-to-BEV perspective transformation presents challenges for precise BEV temporal alignments. This is particularly challenging when aligning a down-scaled BEV feature map, given that traffic lanes occupy only a small portion of the dense BEV space. Consequently, these difficulties might restrict the effectiveness of utilizing historical information for 3D lane detection.
Motivated by StreamPETR~\cite{wang2023exploring} which fuse history query information, we present a novel method for performing temporal fusion of historical results in this study. Our approach involves utilizing history sparse curve queries and dynamic anchor point sets, enabling efficient temporal propagation. As shown in Fig.~\ref{fig:fig_compare_4Dmethods} (b), our temporal fusion method does not rely on the dense BEV feature map.

To verify the performance of the proposed algorithm, we evaluate our CurveFormer++ on the OpenLane dataset~\cite{chen2022persformer} and ONCE-3DLanes dataset~\cite{yan2022once}. Our proposed CurveFormer++ archives competitive results for 3D lane detection on the  ONCE-3DLanes dataset. It also achieves promising performance on the OpenLane dataset compared with recently proposed Transformer-based 3D lane detection approaches. The effectiveness of each component is validated as well.

In general, our main contributions are three-fold:
\begin{itemize}
    \item We propose CurveFormer++, a novel Transformer-based 3D lane detection algorithm, by formulating queries in decoder layers as dynamic anchor point sets, and a curve cross-attention module is applied to compute the query-to-image similarity. Additionally, a dynamic anchor range iteration scheme in decoder layers is applied for accurate feature extraction.

    \item We introduce a lane-centric temporal modeling paradigm where long-term historical information is propagated through sparse queries and anchor points frame by frame. To this end, CurveFormer++ can incorporate temporal information into the 3D lane detection task from an image sequence using selective temporal curve queries and historical anchor points.
    
    \item Experimental results show that our method achieves promising performance compared with both CNN-based and Transformer-based state-of-the-art approaches on several public datasets.
\end{itemize}

\textbf{Differences from the conference paper.} The preliminary version of this work, CurveFormer~\cite{bai2023curveformer} was accepted at ICRA 2023. The enhancements in comparison to the conference version are outlined as follows: 
\begin{itemize}
\item{We extend CurveFormer~\cite{bai2023curveformer} to encode temporal information from an image sequence. Specifically, we propose a Temporal Curve Cross Attention module that enables the model to propagate historical information through sparse curve queries and dynamic anchor point sets frame by frame. The introduced temporal fusion approach enables the stability of the 3D lane prediction results across frames.}
\item{We modify the curve anchor modeling to incorporate an anchor point range restriction that dynamically controls the number of anchor points. Due to that the length of each lane is not fixed, the proposed range restriction allows the anchor point set to effectively extract lane image features. In this way, it ensures that increasing the number of anchor points does not lead to the inclusion of ineffective image features.}
\item{We conduct experiments on a new dataset ONCE-3DLanes~\cite{yan2022once}, and also proivide additional experiments on OpenLane~\cite{chen2022persformer} dataset to evaluate our method. The extended CurveFormer++ demonstrates substantial improvements over the original CurveFormer~\cite{bai2023curveformer} by a significant margin.}
\end{itemize}

\section{Related Work}
\begin{figure*}[htb]
    \centering
    \includegraphics[width=0.9\linewidth]{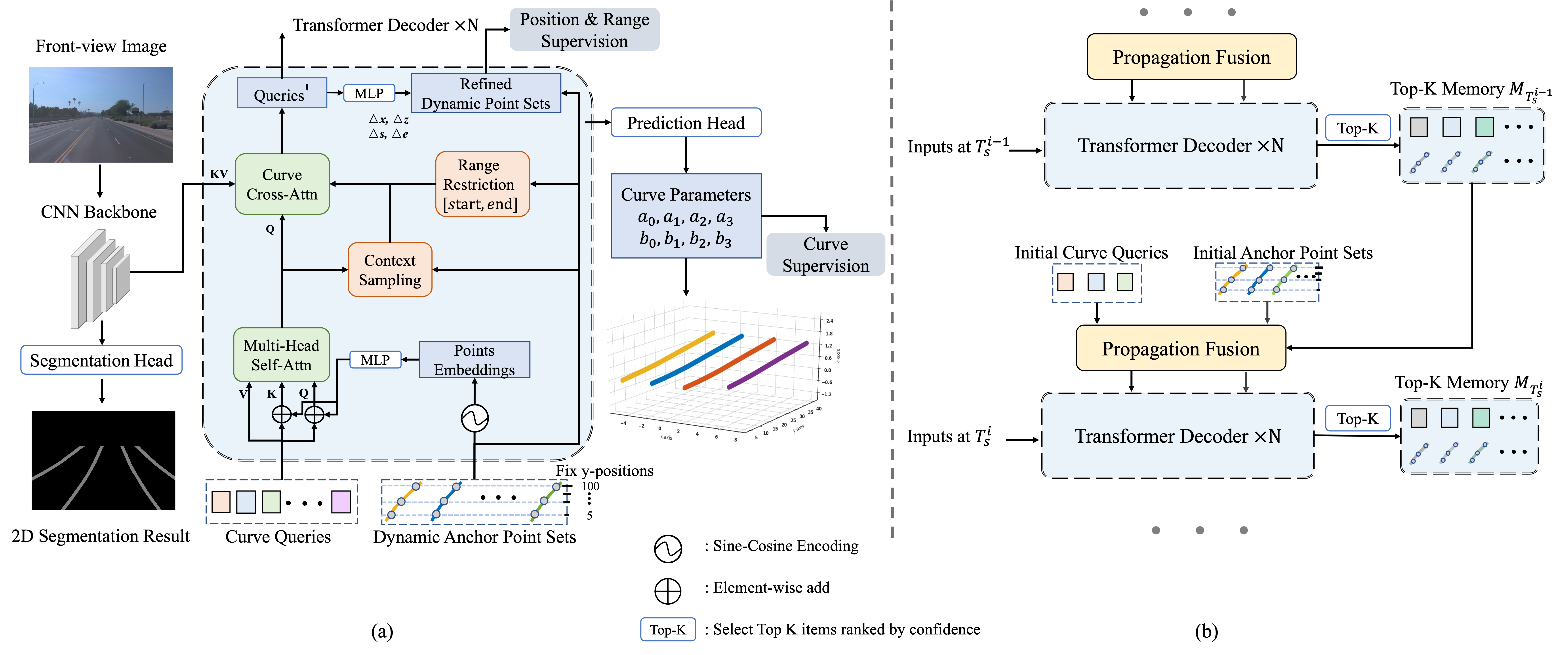}
    \caption{Overview of our proposed CurveFormer++ for single-frame 3D lane detection (a) \&  temporal propagation fusion block in CurveFormer++-T (b).
}
    \label{fig:overview}
\end{figure*}

\noindent\textbf{2D Lane Detection.} Early deep learning lane detection methods are performed in the image plane which can be categorized into segmentation approaches~\cite{pan2018spatial, neven2018towards,hou2019learning,zheng2021resa,10182290,9872124} 
and regression approaches~\cite{chen2019pointlanenet, ko2021key, wang2022keypoint, xu2020curvelane,  li2019line,tabelini2021keep,zheng2022clrnet,qin2020ultra, han2022laneformer,9768189,10196360}.
Segmentation methods differentiate each lane line in an image by assigning a predefined set of semantic labels. 
Approaches have been introduced to capture the structural priors of lanes in addition to standard segmentation methods.
SCNN~\cite{pan2018spatial} first introduces slice-by-slice convolution within feature maps as it is more suitable for lane detection. 
LaneNet~\cite{neven2018towards} upgrades semantic segmentation to instance segmentation by adding an embedding branch for clustering. SAD~\cite{hou2019learning} proposes a pluggable self-attention module to enhance the learning ability of the feature without additional supervision. RESA~\cite{zheng2021resa} combines spatial information in vertical and horizontal directions based on strong shape priors of lanes. Bi-Lanenet~\cite{10182290} proposes a new bilateral lane recognition network using the random sample consensus. MFIALane~\cite{9872124} aggregates multi-scale feature information and employs a channel attention mechanism. 

Instead of using global segmentation methods, recent studies also formulate lane detection to a local lane line regression task.
Lane regression algorithms can be grouped into key points estimation~\cite{ko2021key, wang2022keypoint,9768189,10196360}, anchor-based regression~\cite{chen2019pointlanenet,xu2020curvelane,  li2019line,tabelini2021keep,zheng2022clrnet} and row-wise regression~\cite{qin2020ultra, han2022laneformer}. 
PINet~\cite{ko2021key} formulates lanes by independent key points and uses instance segmentation to solve clustering problems and GANet~\cite{wang2022keypoint} represents lanes as a set of key points which are all related to the start point. DevNet~\cite{9768189} combines deviation awareness with semantic features based on point estimation. Point2Lane~\cite{10196360} selects variable quantity principal points and simply reproduces the target lane by connecting them.
PointLaneNet~\cite{chen2019pointlanenet} and CurveLane-NAS~\cite{xu2020curvelane} divide images into non-overlapping grids and regress lanes offset relative to vertical line proposals.
Line-CNN~\cite{li2019line} and LaneATT~\cite{tabelini2021keep} regresses lanes on the pre-defined ray-proposals, while CLRNet~\cite{zheng2022clrnet} set the starting point and angle of the ray-anchor as learnable parameters and refine between layers of the feature pyramid. 
Ultra-Fast~\cite{qin2020ultra} treats lane detection as a row-wise classification method which significantly reduces the computational cost.
LaneoFormer~\cite{han2022laneformer} reconstructs the conventional transformer architecture by row-column self-attentions to better obtain the shape and semantic information of lanes.

In addition to point regression, modeling lanes using polynomial equations is another approach that has been explored.
PolyLaneNet~\cite{tabelini2021polylanenet} uses global features to directly predict the polynomial coefficients in the image plane. PRNet~\cite{wang2020polynomial} adds two auxiliary branches: initial classification and height regression to enhance polynomial estimation. Method in~\cite{van2019end} applies IPM and least square fitting to directly predict parabolic equations in BEV space. LSTR~\cite{liu2021end} uses Transformer to interact with image features and lane queries to directly predict 3D lane parameters.

\noindent\textbf{3D Lane Detection.} 
In recent years, there has been a growing trend towards adopting end-to-end methods for lane detection, specifically in the context of 3D lane detection. 
Most CNN~\cite{garnett20193d, efrat20203d, efrat2020semi, guo2020gen, luo2022m, wang2023bev, li2023grouplane} and Transformer~\cite{liu2022learning, chen2022persformer} based methods first construct a dense BEV feature map and subsequently extract the 3D lane information from this intermediate representation.
3D-LaneNet~\cite{garnett20193d} proposes a dual-pathway architecture that uses IPM to transpose features and detect lanes by vertical anchor regression. To solve anchor restrictions on the direction of lanes, 3D-LaneNet+~\cite{efrat20203d} divides the BEV features into non-overlapping cells and reformulates lanes by the lateral offset distance, angle, and height offset relative to the cell center. 
Method in~\cite{efrat2020semi} introduces uncertainty estimation to enhance the capabilities of the network of~\cite{garnett20193d}. 
Gen-LaneNet~\cite{guo2020gen} first introduces a virtual top-view coordinate frame for better feature alignment and proposes a two-stage framework that decouples lane segmentation and geometry encoding. BEVLaneDet~\cite{wang2023bev} presents a virtual camera to guarantee spatial consistency and represents 3d lanes by key points to adapt to more complex scenarios.
GroupLane~\cite{li2023grouplane} first introduces a row-wise classification method in BEV which can support lanes in any direction and interact with feature information within instance groups. 
As camera pose estimation is the key to 3D lane detection, CLGo~\cite{liu2022learning} proposes a two-stage framework that estimates camera pose from image and decoder lanes from the BEV feature. PersFormer~\cite{chen2022persformer} builds a dense BEV query with an offline camera pose and unifies 2D and 3D lane detection under the transformer-based framework. WS-3D-Lane~\cite{ai2023ws}  introduces a weakly supervised 3D lane detection method which indirectly supervises the training of 3D lane heights by assuming constant lane width and equal height on adjacent lanes.
STLanes3D~\cite{wang2022spatio} predicts 3D lanes using fused BEV features and introduces 3DLane-IOU loss to couple the errors in the lateral and height directions.
To reduce computational overhead, some methods have recently tried to detect 3D lanes without explicitly building BEV features. 
These methods are based on the flat road assumption. However, this assumption becomes invalid when facing uphill or downhill terrains, which are frequently encountered in driving scenarios. As a result, the latent feature representation is affected by unanticipated distortions due to variations in road elevation, reducing the model's reliability and posing risks to traffic safety.

Anchor3DLane~\cite{huang2023anchor3dlane}, a CNN-based method that regresses 3D lanes directly from image features based on 3D anchor. 
Nevertheless, this BEV-free approach necessitates a dense anchor design to alleviate perspective geometric distortion, which can result in complex and unclear lane representations.
For example, CurveFormer~\cite{bai2023curveformer} utilizes the sparse query representation and cross-attention mechanisms in Transformer to efficiently regress the 3D lane polynomial coefficient.  Following similar query anchor modeling of CurveFormer~\cite{bai2023curveformer}, LATR~\cite{luo2023latr} constructs a lane-aware query generator and dynamic 3D ground positional embedding to extract the lane information.

In addition to using the camera as sensor input, $\textrm{M}^{2}$-3DLaneNet~\cite{luo2022m} incorporates LiDAR data to enhance monocular 3D lane detection by lifting image features into 3D space and then fuse the multi-modal features in the BEV space. DV-3DLans~\cite{luodv} combines image and LiDAR data to improve accuracy through effective multi-modal feature learning and dual-view deformable attention mechanisms.

\noindent\textbf{Temporal Fusion.} 
Temporal information from history frames provides additional information for 3D perception in autonomous driving. 
For example, in 3D object detection tasks, BEVFormer~\cite{li2022bevformer} introduces temporal modeling into multi-view 3D object detection, it employs temporal self-attention to fuse the historical BEV features.
Sparse4D~\cite{lin2022sparse4d} iteratively refines 3D anchor by sparse sampling and fusing multi-dimensional features in a specific order to obtain accurate detection results.
StreamPETR~\cite{wang2023exploring} feds both previous and current sparse object queries into a Transformer decoder to perform the spatial-temporal interaction.

In 3D lane detection tasks, Anchor3DLane-T~\cite{huang2023anchor3dlane} incorporates temporal information by projecting the 3D anchors of the current frame onto previous frames to sample features.
PETRv2~\cite{liu2023petrv2} extends the 3D position embedding and multi-view image features for temporal modeling to provide informative guidance for the query learning in the Transformer decoder. 
STLane3D~\cite{wang2022spatio} proposes a novel multi-frame pre-alignment layer under the BEV space, which uniformly projects features from different frames onto the same ROI region. Although these methods aim to acquire historical information from either front-view features or Bird's Eye View (BEV) features, the issue of misalignment remains inevitable.

\section{Method}
\subsection{Overview}




Fig.~\ref{fig:overview} (a) and (b) show the overview of our CurveFormer++ for single frame 3D lane detection and the proposed temporal fuse block additionally used in CurveFormer++-T, respectively. 

In Fig.~\ref{fig:overview} (a), CurveFormer++ consists of two major components: a Shared CNN Backbone takes a single front-view image as input and outputs multi-scale feature maps, and a Curve Transformer Decoder to propagates curve queries by curve cross-attention and iteratively refine anchor point sets. Finally, a prediction head is applied to output 3D lane parameters. The $i$-th output can be represented as $\text{Pred}_i = (p_i, y_i^{start}, y_i^{end}, \{a_i, b_i\}_{r=0}^R)$,
where $p_i$ is the foreground confidence, $y_i^{start}$ and $y_i^{end}$ are start and end point in the $Y$ direction. Two polynomials of the 3D lane are denoted by $a_i$ and $b_i$ with order $R$ to model a traffic lane in X-O-Y and Y-O-Z plane, respectively.

Based on the CurveFormer++ framework, we design a temporal fusion block that leverages the sparse output from the preceding time step as prior information for the current time step. In Fig.~\ref{fig:overview}. (b), we specifically highlight the distinction between CurveFormer++-T and CurveFormer++ within the Transformer decoder module, while maintaining consistency in the design of other network structures. Our novel temporal propagation fuse block incorporates sparse curve queries and dynamic anchor point sets, enabling the CurveFormer++-T Transformer decoder to retrieve the Top-K outputs ranked by confidence score from the previous time step and integrate them with the initial curve queries and anchor point sets from the current time step.

\subsection{Shared CNN Backbone}
The backbone takes an input image and outputs multi-scale feature maps. We add an auxiliary segmentation branch in the training stage to enhance the shared CNN backbone.

\subsection{Sparse Curve Query with Dynamic Anchor Point Set}
\label{subsec:Representing}

DAB-DETR~\cite{liu2022dab} introduces a novel approach where queries are modeled as anchor boxes, represented by 4D coordinates (x, y, w, h). This representation allows the cross-attention module to utilize both the position and size information of each anchor box. Taking inspiration from DAB-DETR, we apply a similar approach to the Transformer-based 3D lane detection with dynamic anchor point sets. As shown in Fig.~\ref{fig:representation} (a), these points are sampled on the X-O-Y plane and Z-O-Y plane at a set of fixed $Y$ locations. Typically, we denote $C_i = \{p_1=(x_1,y_1,z_1), \cdots, p_N=(x_N, y_N, z_N)\}$ as the $i$-th anchor point set. Its corresponding content part and positional part are $Z_i \in \mathbb{R}^D$ and $P_i \in \mathbb{R}^D$, respectively. The positional query $P_i$ is calculated by:
\begin{equation}
    \text{P}_i = \text{MLP}(\text{PE}(C_i))
\label{P_i}
\end{equation}
\begin{equation}
    \text{PE}(C_i) = \text{Concat}(\text{PE}(\{x_i\}_1^N), \text{PE}(\{y_i\}_1^N), \text{PE}(\{z_i\}_1^N)),
\label{P_i}
\end{equation}
where positional encoding (PE) is utilized to generate embeddings using floating numbers, while the parameters of the MLP are shared across all layers.


\begin{figure}[t!]
 \centering
  \begin{tabular}{c}
   \includegraphics[width=0.9\linewidth]{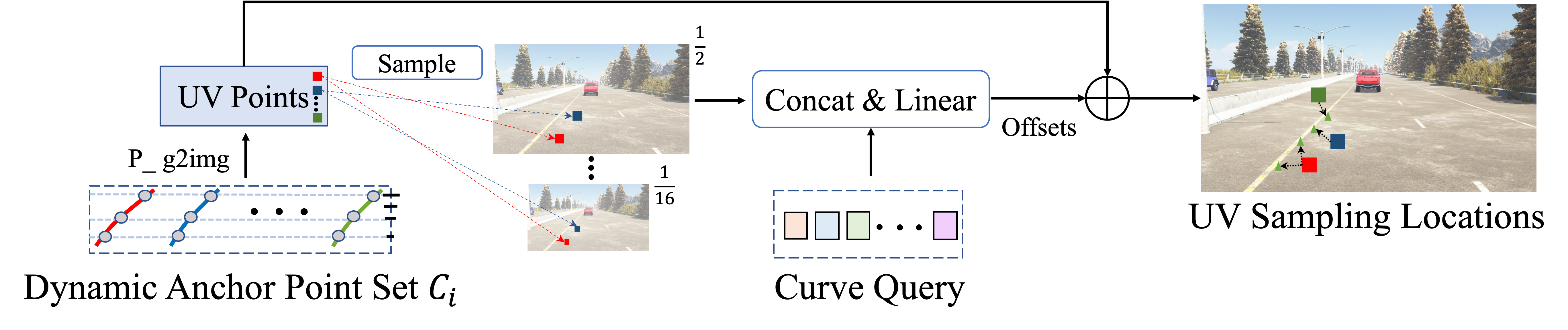}\\
 \end{tabular}
\caption{Illustration of the Context Sampling Module. Our context sampling module learns sampling offsets by leveraging both queries and image features.}
\label{fig:context-sampling}
\end{figure}

\begin{figure*}[ht]
	\centering
	\includegraphics[width=0.99\textwidth, height=3.8cm]{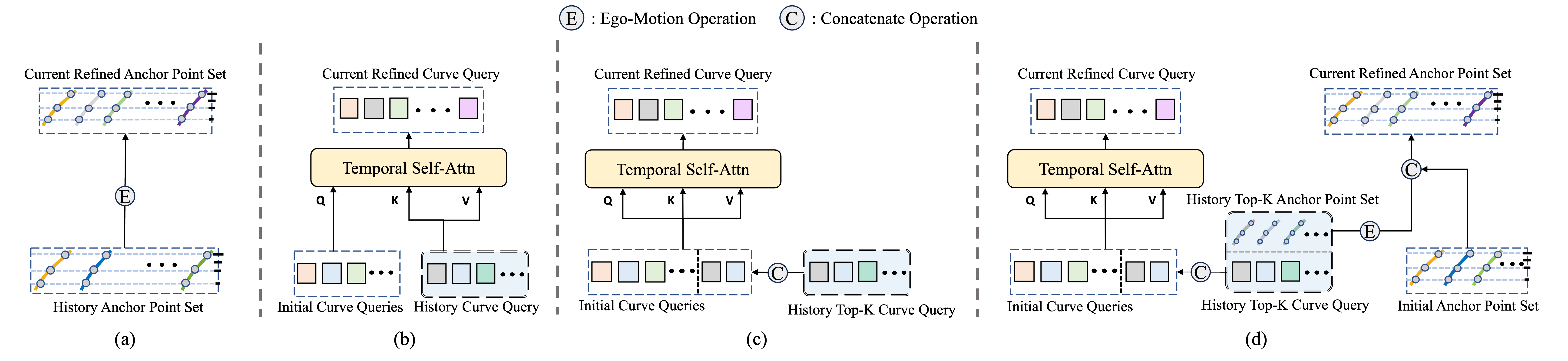}
	\caption{The details of four temporal fusion modules. (a) Utilizing historical anchor point set. (b) Utilizing historical curve query. (c) Utilizing historical Top-K curve query. (d) Utilizing historical Top-K curve query \& anchor point set. 
    }
	\label{fig:compare_temporal_methods}
\end{figure*}

By representing a curve query as an ordered anchor point set $\{p_1 \dots p_N\}$, we can iteratively refine the curve query in the Transformer decoder. Specifically, each Transformer decoder layer estimates relative positions $(\{\Delta x\}_1^N, \{\Delta z\}_1^N)$ by a shared parameters linear layer. In this way, the curve query representation is suitable for 3D lane detection and enhances learning convergence by employing a layer-by-layer refinement scheme. Fig.~\ref{fig:representation} (b) demonstrates the refinement process in the image space, showing how the dynamic anchor point set is progressively refined by the decoder, gradually converging to the ground truth with each iteration.

\subsection{Curve Transformer Decoder}

Our curve Transformer decoder consists of three main components: a multi-head self-attention module, a context sampling module and a curve cross-attention module. We apply deformable attention~\cite{zhu2020deformable} in the self-attention module which focuses on a limited set of key sampling points surrounding the reference point, regardless of the spatial size of the feature map.

\noindent\textbf{Context Sampling Module.} In the deformable DETR~\cite{zhu2020deformable} approach, a learnable linear layer is employed to estimate the offsets of sampling locations associated with the reference points by queries, which are irrelevant to the image features. In contrast to this method, we introduce a context sampling module that predicts sampling offsets by incorporating more relative image features. 
Fig.~\ref{fig:context-sampling} illustrates the difference between the standard sample offset module (a) and our context sampling module (b). 

Firstly, a dynamic anchor point set $C_i$ is projected to the image view using camera parameters. We apply bilinear interpolation to extract features from these projected points $C_i^{2D} = \{p_1^{2D}=(u_1^i,v_1^i), \cdots, p_N^{2D}=(u_N^i, v_N^i)\}$ on multiscale feature maps ${\bX}$. 
The final feature $f_{C_i}$ is computed by
\begin{equation}
  f_{C_i} = \frac{1}{\sum_{l=1}^{L} \sum_{n=1}^{N}\sigma_{ln} + \epsilon} \sum_{l=1}^{L} \sum_{n=1}^{N}    \bX_{l}(p_n^{2D}) \sigma_{ln}, 
\end{equation}
where $\sigma_{ln}$ is used to determine whether a projected point $p_n^{2D}$ is outside $l$-th feature map. And $\epsilon$ is a small number to avoid division by zero.

Afterward, we employ a trainable linear layer to forecast $K$ sampling offsets. Generally, for a curve query $\bZ_q$ having an anchor point set $C_i$, the context sampling module is represented as:
\begin{align}
    \text{CS}(\{\Delta u_{nk}^i, \Delta v_{nk}^i\}) = \text{MLP} (\text{Concat}(f_{C_i}, \bZ_q)),
\label{context-sampling}
\end{align}
where $n=1,\cdots, N$ and $k=1,\cdots, K$.

\noindent\textbf{Anchor Point Range Restriction.} In real-world scenarios, the lengths of lanes can vary, making it impractical to rely on a sparse fixed-$Y$ anchor point set to accurately represent a lane. Also, it is unlikely for the actual lane length to precisely match the sampled length, resulting in insufficient ground truth supervision for the anchor points. To address this challenge, we propose an anchor point range restriction module that predicts the starting and ending positions of each anchor point set.
Similar to estimating the relative positions of each anchor point in ~\ref{subsec:Representing}, each decoder layer also predicts the anchor point ranges $(\Delta s, \Delta e)$ for every anchor point set $C_i$ using a shared parameter linear layer. 
In addition, unlike previous methods~\cite{chen2022persformer, bai2023curveformer, guo2020gen}, we employ denser sampling along the $Y$-axis to construct a dynamic anchor point set using anchor point range restriction. To this end, our model allows for the dynamic adjustment of anchor point sets to enhance the robustness and adaptability of the model in accurately representing lanes with varying lengths.

\noindent\textbf{Curve Cross Attention.} We incorporate the deformable attention module from Deformable DETR~\cite{zhu2020deformable} into our curve cross-attention module. 
Mathematically, considering a query element $q$ within $\bZ_q$ and its corresponding anchor point set $C_i$, the computation of our curve cross-attention can be expressed as:
\begin{align}
     \text{CCA}:\left(\bZ_q, C_i,\left\{\bx^l\right\}_{l=1}^L\right)= 
     \sum_{m=1}^M \bW_m \notag \\
     \left[\sum_{l=1}^L \sum_{n=1}^N A_{m l n} \cdot \bW_m^{\prime} \bx^l\left(\phi_c (p_n)+\Delta \bp_{m l  n}\right)\right],
\label{cross-attn}
\end{align}
where $(m,l,n)$ represents the indices of the attention head, feature level, and the sampling point. $\Delta \bp_{m l n}$ and $A_{m l n}$ denote sampling offsets and attention weights of the $n$-th sampling point in the $l$-th feature level and the $m$-th attention head. The scalar attention weight $A_{m l n}$ is normalized to sum as 1. The function $\phi_c(\cdot)$ is responsible for rescaling the normalized coordinates to align with the input feature maps.

\subsection{Propagation Fusion Module}

In autonomous driving, static lane instances observed in the current frame tend to persist in subsequent frames. This observation motivates us to propagate historical information to the following frame. Based on the unique design of CurveFormer++, we propose a straightforward temporal fusion module that incorporates curve query and its corresponding anchor point set. As illustrated in Fig.~\ref{fig:compare_temporal_methods}, we compare four different temporal propagation fusion modules that build upon CurveFormer++ to validate the improvement contributed by fusing historical information.

\textbf{Utilizing historical anchor point set.} As shown in Fig.~\ref{fig:compare_temporal_methods} (a), the simplest way is to replace the initial anchor point set of the current frame with the one from historical frames using ego-motion information. Typically, given a 3D point $(x_{t-i}, y_{t-i}, z_{t-i})$ of ground coordinate at $t-i$, we can transform it to ground coordinate of $t$-th frame: 
\begin{equation}
\left[\begin{array}{c}
x_{t} \\
y_{t} \\
z_{t} \\
1
\end{array}\right]=\mathbf{T}_{g(t-i) \rightarrow g(t)}\left[\begin{array}{c}
x_{t-i} \\
y_{t-i} \\
z_{t-i} \\
1
\end{array}\right],
\label{equ:3}
\end{equation}
where $\mathbf{T}_{g(t-i) \rightarrow g(t)}$ denotes the transformation matrix from $(t-i)$-th frame to $t$-th frame.

\textbf{Utilizing historical curve query.} 
In contrast to BEVFormer~\cite{li2022bevformer}, which incorporates temporal modeling into multi-view 3D object detection using self-attention to fuse historical BEV queries, we propose an alternative framework to employ sparse queries as hidden states for temporal propagation. Specifically, we transfer historical information from sparse curve queries to current frames. This is accomplished by utilizing historical frame curve queries as both \textit{keys} and \textit{values}, and applying temporal self-attention to the initial curve queries of the current frame, shown in Fig.~\ref{fig:compare_temporal_methods} (b).

\textbf{Utilizing historical Top-K curve query.} In Transformer-based detection models, the number of queries typically far exceeds the actual number of targets, leading to a situation in which some queries are unable to learn the representation of targets effectively. Following ~\cite{wang2023exploring, yuan2024streammapnet}, we only propagate the curve queries with the highest confidence scores to the next frame. As shown in Fig.~\ref{fig:compare_temporal_methods} (c), we directly concatenated historical Top-k curve queries into the current randomly initialized queries and fuse curve queries as the \textit{queries}, \textit{keys}, and \textit{values} with temporal self-attention.

\textbf{Utilizing historical Top-K curve query \& anchor point set.} To fully leverage the spatial and contextual priors, we extend the propagation beyond the Top-K curve queries by also transferring the corresponding sets of anchor points to the current frame. As shown in Fig.~\ref{fig:compare_temporal_methods} (d), when propagating the historical Top-K curve queries, we transform the Top-K anchor point sets to the current coordinate system using equation (\ref{equ:3}), and then concatenate them with the initial anchor point sets. By including both the historical curve queries and their associated anchor points, the model can capture and exploit the temporal dependencies and spatial relationships between lanes, leading to improved performance of 3D lane detection.

\subsection{Curve Training Supervision}

This section outlines the training supervision for our model. Apart from the enhanced anchor point set $\bP = {p_n}_{n=1}^N$, our model's prediction head also generates curve parameters for $L$ 3D lanes, where $L$ exceeds the maximum number of labeled lanes found in the training dataset.
Similarly to~\cite{liu2022learning}, we first associate the predicted curves $\text{Pred}_i = (p_i, y_i^{start}, y_i^{end}, \{a_i, b_i\}_{r=0}^R)$ and the ground truth lanes $\text{GT}_i = (\hat{p}_i, \hat{y}_i^{start}, \hat{y}_i^{end}, \hat{\bL}_i=\{\hat{p}_n\}_1^N)$ by solving a bipartite matching problem, where $c \in\{0,1\}$ (0: background, 1: lane). We sample a set of 3D points $\bL_i=\{p_n\}_1^N$ using the predicted curve parameters to calculate the loss of matching and training. The boundary of the lane (start and end points) is indicated by $\bL_i^b=\{ y_i^{start}, y_i^{end} \}$.

Let $\Omega = \left\{ w_l = \text{Pred}_l \right\}_{l=1}^L$ be the set of predicted 3D lanes and $\Pi = \left\{ \hat{\pi}_l = \text{GT}_l \right\}_{l=1}^L$ be the set of ground truth.
Note that $\Pi$ is padded with non-lanes to fill enough the number of ground truth lanes to $L$. The matching problem is formulated as a cost minimization problem by searching an optimal injective function $z: \Pi \rightarrow \Omega$, where $z(l)$ is the index of a 3D lane prediction $\omega_{z(l)}$ which is assigned to $l$-th ground truth 3D lane $\hat{\pi}_l$:
\begin{equation}
    \hat{z}=\underset{z}{\arg \min } \sum_{l=1}^L D\left(\hat{\pi}_l, \omega_{z(l)}\right).
\label{matching-function}
\end{equation}
The matching cost is calculated as:
\begin{align}
    D=-\alpha_1 p_{z(l)}\left(\hat{c}_l\right)+\mathds{1}\left(\hat{c}_l=1\right) \alpha_2 \left|\hat{\bL}_l-\bL_{z(l)}\right| \notag \\
    +\mathds{1}\left(\hat{c}_l=1\right) \alpha_3 \left|\hat{\bL}_l^b-\bL_{z(l)}^{b}\right|,
\label{matching-cost}
\end{align}
where $\alpha_1$, $\alpha_2$, and $\alpha_3$ are coefficients that adjust the loss effects of classification, polynomial fitting, and boundary regression, and $\mathds{1}$ is an indicator function. 

After solving Eq.~\ref{matching-function} by Hungarian algorithms~\cite{carion2020end}, the final training loss can be written as:
\begin{equation}
L_{total}=L_{curve} + L_{query} + L_{seg},
\end{equation}
where $L_{curve}$ is the curve prediction loss, $L_{query}$ is the deep supervision of refined anchor point set for each curve, and $L_{seg}$ is an auxiliary segmentation loss.
The curve prediction loss is defined as:
\begin{align}
    L_{curve}= \sum_{l=1}^L -\alpha_1 \log p_{\hat{z}(l)}\left(\hat{c}_l\right) + \mathds{1}\left(\hat{c}_l=1\right) \notag \\ \alpha_2\left|\hat{\bL}_l-\bL_{\hat{z}(l)}\right| 
    + \mathds{1}\left(\hat{c}_l=1\right) \alpha_3 \left|\hat{\bL}_l^b-\bL_{z(l)}^{b}\right|,
\label{3D-loss}
\end{align}
where $\alpha_1$, $\alpha_2$, and $\alpha_3$ are the same coefficients with Eq.~\ref{matching-cost}, and deep supervision of refined anchor point set including anchor point position and range of a lane:
\begin{align}
    L_{query} = \sum_{l=1}^L \mathds{1}\left(\hat{c}_l=1\right) \alpha_4\left|\hat{\bL}_l - \bP_{\hat{z}(l)}\right| \notag \\ 
    + \mathds{1}\left(\hat{c}_l=1\right) \alpha_5 \left|\hat{\bL}_l^b-\bL_{z(l)}^{b}\right|.
\label{point-loss}
\end{align}

\section{EXPERIMENTS}

\begin{table*}[t]
  \centering
  \caption{Comprehensive 3D Lane evaluation on ONCE-3DLanes and OpenLane datasets. The best results are highlighted in bold, and the second best are underlined (C and L represent camera and LiDAR input).}
  \label{tab:results:all}
  \begin{tabular}{cccccccccc}
    \toprule
     & \multirow{2}{*}{Method} & \multirow{2}{*}{Venue} & \multirow{2}{*}{Modality} & Dense & Lane & PE & ONCE & OpenLane & OpenLane-300\\
     &  & & & BEV & Embed & Embed & F1(\%) & F1(\%) & F1(\%)\\
    \midrule
    \multirow{5}{*}{CNN} & 3D-LaneNet\cite{garnett20193d} & ICCV 2019 & C &  \checkmark & -  & - & 44.73 & 44.1 & -  \\
    & Gen-LaneNet\cite{guo2020gen}  & ECCV 2020 & C & \checkmark & -  & - & 45.59 & 32.3 & - \\
    & WS-3D-Lane\cite{ai2023ws} & ICRA 2023 & C & \checkmark & -  & \checkmark & 77.02 & - & - \\
    & $\textrm{M}^2$-3DLaneNet\cite{luo2022m} & Arxiv 2023 & C+L & \checkmark & - & - & - & \underline{62.0} & -\\
    & Anchor3DLane\cite{huang2023anchor3dlane} & CVPR 2023 & C & - & -  & - & 74.44 & 53.7 & - \\

    \midrule
    \multirow{8}{*}{Transformer} & PersFormer\cite{chen2022persformer} & ECCV 2022 & C & \checkmark & - & - & 72.07 & 50.5 & -  \\  
    & STLane3D\cite{wang2022spatio} & BMVC 2022 & C & \checkmark & -  & - & 74.05 & - & -\\
    & STLane3D-T\cite{wang2022spatio} & BMVC 2022 & C & \checkmark & - & - & 77.55 & 50.6 & - \\
    & LATR\cite{luo2023latr} & ICCV 2023 & C & - & - & - & - & - & 45.5\\
    & LATR\cite{luo2023latr} & ICCV 2023  & C & - & \checkmark & \checkmark & \textbf{80.59} & 61.9 & \textbf{70.4}\\
    & DV-3DLane\cite{luodv} & ICLR 2024 & C+L & \checkmark &  - & - & - & \textbf{66.8} & - \\
    \cline{2-10}
    \rule{0pt}{8pt}
    & CurveFormer++ & - & C &  - & - & - & 77.22 & 52.7 & 54.3\\
    & CurveFormer++-T & - & C & - & - & - & \underline{77.85} & 52.5 & \underline{55.4}\\
    \bottomrule 
  \end{tabular}
\end{table*}

\begin{table*}[t]
  \begin{center}
    \caption{Comprehensive 3D Lane evaluation on ONCE-3DLanes dataset. The best results are highlighted in bold, second best are underlined.}
    \label{tab:results-once-val-set}
    \begin{tabular}{ccccccc}
    \toprule
      & Method & Dense BEV & F1(\%) & Precision(\%) & Recall(\%) & CD error(m) \\
      \midrule
      \multirow{3}{*}{CNN} 
      & 3D-LaneNet\cite{garnett20193d} & \checkmark & 44.73 & 61.46 & 35.16 & 0.127 \\
      & Gen-LaneNet\cite{guo2020gen} & \checkmark & 45.59 & 63.95 & 35.42 & 0.121 \\
      & Anchor3DLane\cite{huang2023anchor3dlane} & - & 74.44 & 80.50 & 69.23 & \textbf{0.064} \\
      \midrule
      \multirow{5}{*}{Transformer} 
      & PersFormer\cite{chen2022persformer} & \checkmark & 72.07 & 77.82 & 67.11 & 0.086 \\
      & STLane3D\cite{wang2022spatio} & \checkmark & 74.05 & 76.63 & 71.64 & 0.085 \\
      & STLane3D-T\cite{wang2022spatio} & \checkmark & \underline{77.55} & \underline{81.54} & \underline{73.90} & \underline{0.066} \\
      \cline{2-7}
      \rule{0pt}{8pt}
      & CurveFormer++ & - &77.22 &\textbf{82.22} &72.79 &0.081 \\
      & CurveFormer++-T & - &\textbf{77.85} &81.06 &\textbf{74.89} & 0.084 \\
      \bottomrule 
    \end{tabular}
  \end{center}
\end{table*}

\begin{table*}[t]
  \begin{center}
    \caption{Performance comparison with other state-of-the-art 3D lane methods on OpenLane benchmark at different scenario sets. The best results are highlighted in bold, the second best are underlined, third best are in italics.} 
    \label{tab:results-openlane-scenario-set}
    \begin{tabular}{cccccccccc}
    \toprule
      Method & Dense BEV & All & Up\&Down & Curve & Extreme Weather & Night & Intersection & Merge\&Split & Mean $F_{\text{stab}}$\\
      \midrule
      3D-LaneNet\cite{garnett20193d} & \checkmark & 44.1 & 40.8 & 46.5 & 47.5 & 41.5 & 32.1 & 41.7 & -\\
      Gen-LaneNet\cite{guo2020gen} &\checkmark & 32.3 & 25.4 & 33.5 & 28.1 & 18.7 & 21.4 & 31.0 & -\\
      PersFormer\cite{chen2022persformer} & \checkmark & 50.5 & 42.4 & 55.6 & 48.6 & 46.6 & 40.0 & \textit{50.7} & \textit{0.829}\\
      Anchor3DLane\cite{huang2023anchor3dlane} & - & \underline{53.7} & 46.7 & 57.2 & \underline{52.5} & 47.8 & \textit{45.4} &\underline{51.2} & -\\
      Anchor3DLane-T\cite{huang2023anchor3dlane} & - & \textbf{54.3} & \textit{47.2} & \textit{58.0} & \textbf{52.7} & \underline{48.7} & \underline{45.8} & \textbf{51.7}& - \\
      \hline
      \rule{0pt}{8pt}
      CurveFormer ~\cite{bai2023curveformer} & - & 50.5 & 45.2 & 56.6 & 49.7 & \textbf{49.1} & 42.9 & 45.4& -\\
      CurveFormer++ & - &  \textit{52.7}&  \textbf{48.3} &  \textbf{59.4}&  50.6&  48.4&  45.0& 48.1& \underline{0.227}\\
      CurveFormer++-T & - & 52.5&  \underline{48.1}&  \underline{59.0}&  \textit{51.2}&  \textit{48.5}&  \textbf{51.2}&  48.1 & \textbf{0.215} \\
      \bottomrule
    \end{tabular}
  \end{center}
\end{table*}

\begin{table*}[t]
  \begin{center}
    \caption{Comprehensive 3D Lane evaluation on OpenLane dataset.}
    \label{tab:results-openlane-val-set}
    \begin{tabular}{cccccccc}
    \toprule
      Method & F1(\%) & Cate-Acc & X error near & X error far & Z error near & Z error far & Mean $F_{\text{stab}}$ \\
      \midrule
      3D-LaneNet\cite{garnett20193d} & 44.1 & - & 0.479 & 0.572 & 0.367 & 0.443 & - \\
      Gen-LaneNet\cite{guo2020gen} & 32.3 & - &  0.591 & 0.684 & 0.411 & 0.521 & - \\
      Cond-IPM & 36.6 & - & 0.563 & 1.080 & 0.421 & 0.892 & -\\
      PersFormer\cite{chen2022persformer} & 50.5 & 89.5 & 0.485 &  0.553 & 0.364 &  0.431 &  0.829\\
      Anchor3DLane\cite{huang2023anchor3dlane}  & 53.7 & \textbf{90.9} & 0.276 & 0.311 & 0.107 & 0.138 & - \\
      Anchor3DLane-T\cite{huang2023anchor3dlane}  & 54.3 & 90.7 & 0.275 & \textbf{0.310} & 0.105 & 0.135 & -\\
      BEVLaneDet\cite{wang2023bev} & 58.4 & - & 0.309 & 0.659 & 0.244 & 0.631 & -\\
      \hline
      \rule{0pt}{8pt}
      CurveFormer~\cite{bai2023curveformer} & 50.5 & - & 0.340 & 0.772 &  0.207 & 0.651 & -\\
      CurveFormer++  &52.7& 88.1 & 0.337& 0.801& 0.198& 0.676 & 0.227\\
      CurveFormer++-T &52.5 &87.8 & 0.333 & 0.805 & 0.186 & 0.687 & \textbf{0.215}\\
      \bottomrule 
    \end{tabular}
  \end{center}
\end{table*}

\subsection{Dataset}

\textbf{OpenLane Dataset.} OpenLane dataset~\cite{chen2022persformer} is the first real-world 3D lane dataset which consists of over 200K frames at a frequency of 10 FPS based on Waymo Open dataset\cite{Sun_2020_CVPR, llc2019waymo}. In total, it has a training set with 157K images and a validation set of 39K images. The dataset provides camera intrinsics and extrinsic parameters following the same data format as Waymo Open Dataset. 

\textbf{ONCE\_3DLanes Dataset.} ONCE\_3DLanes Dataset ~\cite{yan2022once} is a real-world 3D lane dataset constructed from the recent large-scale autonomous driving dataset ONCE~\cite{mao2021one}. The dataset comprises 211K images with high-quality 3D lane point annotations, diverse weather and region conditions.

\subsection{Experiment Settings}
\textbf{Implementation Details.}  We use EfficientNet~\cite{tan2019efficientnet} as the backbone which gives 4 scale feature maps. The input image is resized to $720 \times 960$. The 3D-space range is set to $[-30m,30m] \times [3m,103m] \times [-10m,10m]$ along $x, y$ and $z$ axis respectively. 
For curve representation, we use the fixed $40$ y positions in a uniformly sampled range. We set coefficients to $\alpha_1=2$, $\alpha_2=5$, $\alpha_3=2$, $\alpha_4=2$, and $\alpha_5=2$. All experiments are performed with known camera poses and intrinsic parameters provided by two datasets. Our network uses Adam optimizer~\cite{kingma2014adam}, with a base learning rate of $2 \times 10^{-4}$ and a weight decay of $10^{-4}$. All models are trained from scratch for 50 epochs with 4 NVIDIA RTX 3090 GPUs. The batch size for OpenLane datasets is set to 2, while the ONCE-3DLanes dataset is set to 8.

\subsection{Evaluation Metrics}
\label{subsec:metrics}
\textbf{Existing Evaluation metrics.} For OpenLane datasets, we follow the evaluation metrics designed by Gen-LaneNet~\cite{guo2020gen}. Point-wise Euclidean distance is calculated when a $y$-position is covered by both prediction and the ground-truth. 
For each predicted lane, we consider it matched when $ 75\% $ of its covered $y$-positions have point-wise Euclidean distance less than the max-allowed distance (1.5 meters). We report Average Precision (AP), F-Score, and errors (near range and far range) to investigate the performance of our model. 
The ONCE-3DLanes dataset employs a two-stage evaluation metric for lane detection. Initially, the matching degree is determined based on the Intersection over Union (IoU) on the top-view plane. If the IoU of matched pairs exceeds a certain threshold, they are further evaluated using the unilateral Chamfer Distance (CD) as the matching error. In our evaluation, we report the results in terms of F1 score, precision, recall, and CD error on the ONCE-3DLanes dataset. 

In recent years, significant progress has been made in 4D object detection ~\cite{lin2022sparse4d, wang2023exploring, li2022bevformer, liu2023petrv2} by leveraging various types of temporal information. The fusion of historical information has been shown to enhance the performance of models in occluded scenes. However, the benefits of applying temporal modules in the context of lane detection tasks may not be readily apparent. This is because detecting distant lanes poses a significant challenge, while historical information can only cover nearby regions in the current frame.
Therefore, we introduce a technique to assess the stability of prediction outcomes within a video. Specifically, we evaluate the discrepancy  on the $X$-axis between the predicted results and the ground truth in each frame of the video, subsequently determining the standard deviation of these disparities. The stability metrics for a video segment containing $K$ frames can be established as follows, where $L$ denotes the number of lane markings in each frame and $N$ signifies the number of points in each lane:
\begin{align}
     dist^x = \frac{1}{L \times N} \sum_{i,j}^{L,N} ||\hat p_{ij}^x - p_{ij}^x|| \\
     F_{\text{stab}} = \sqrt{\frac{1}{K}\sum_{k=1}^{K} (dist^x_k - \overline{dist^x} )^2}
\label{stability_metrics}
\end{align}
Finally, the mean value of $F_{\text{stab}}$ is reported as the statistical stability performance over the testing data.

\subsection{Main Results}

Experimental results of F1 score on both ONCE-3DLanes dataset and OpenLane dataset are listed in Table~\ref{tab:results:all}. 
Following a similar query anchor modeling approach as CurveFormer~\cite{bai2023curveformer}, LATR~\cite{luo2023latr} constructs a lane-aware query generator by utilizing semantic results as inputs for the transformer query (Lane-Embed) and dynamic 3D ground positional embedding (Ground-Embed) to enhance lane information. LATR obtains state-of-the-art results on both the OpenLane and ONCE-3DLanes datasets using solely camera input. In the OpenLane subset dataset (($\sim$300 video segments)), our method achieves superior results compared to LATR when Lane-Embed and Ground-Embed components are not utilized. 
As claimed in~\cite{luo2023latr}, the main improvements are achieved by Lane-Embed and Ground-Embed. As our method does not utilize these two components, we primarily compare other approaches that also exclude these components in the subsequent tables.

\begin{figure}[t!]
	\centering
	\includegraphics[width=1\linewidth]{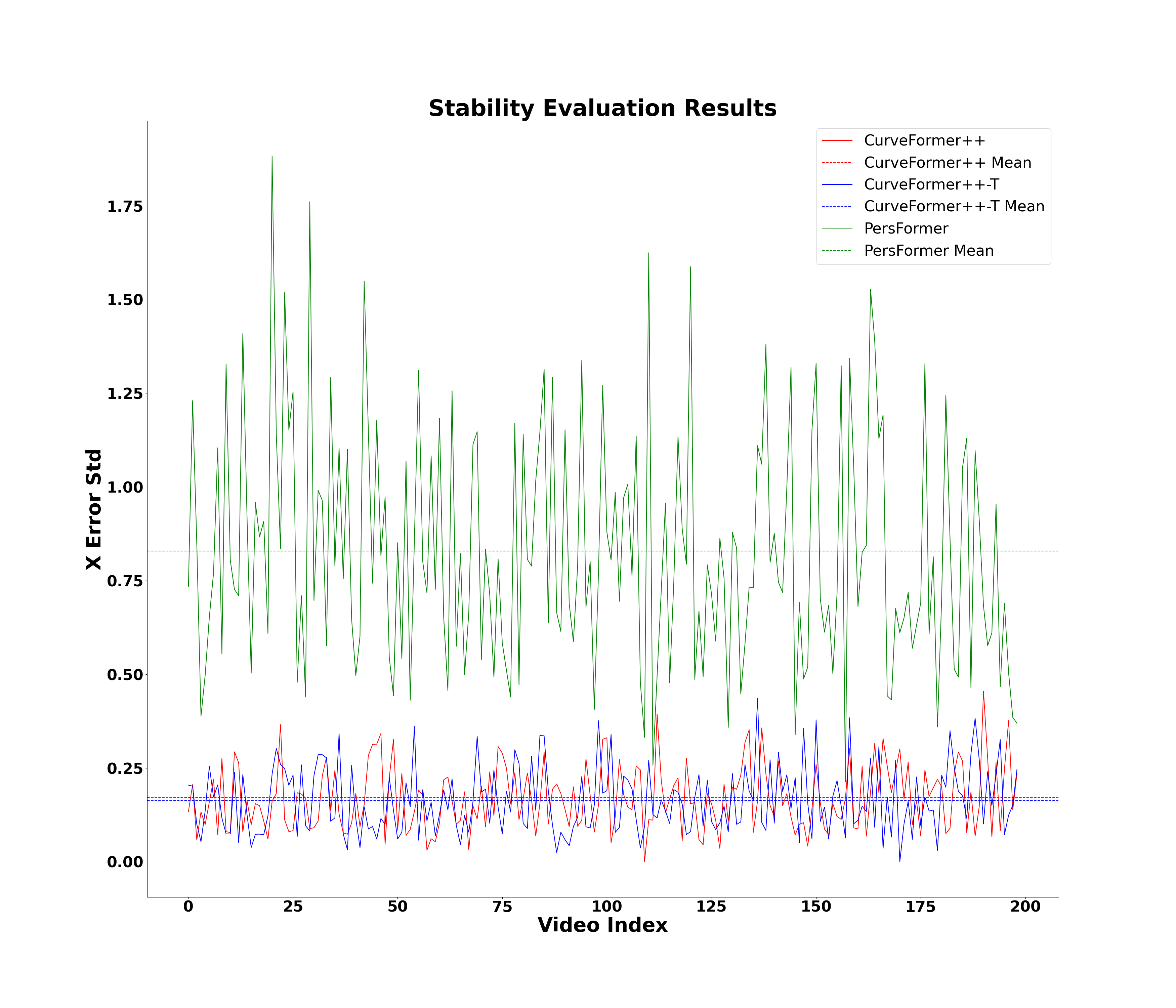}
	\caption{The stability evaluation results of PersFormer~\cite{chen2022persformer}, CurveFormer++, and CurveFormer++-T.}
	\label{fig:video_eval_std}
\end{figure}

\textbf{Results on ONCE-3DLanes Dataset.} Table~\ref{tab:results-once-val-set} presents the experimental results on the ONCE-3DLanes dataset.
Our Transformer-based method outperforms CNN-based methods. For example, CurveFormer++ demonstrates a significant improvement over 3D-LaneNet~\cite{efrat20203d} and Gen-LaneNet~\cite{guo2020gen}. It also outperforms Anchor3DLane~\cite{huang2023anchor3dlane} by 2.78\% in F-Score and 1.72\% in Precision. Furthermore, CurveFormer++ surpasses Transformer-based methods, outperforming PersFormer~\cite{chen2022persformer} by 5.15\% in F-Score and STLane3D~\cite{wang2022spatio} by 3.17\% in F-Score. Notably, CurveFormer++ achieves performance comparable to STLane3D-T, which additionally incorporates temporal feature information. 
Our temporal model, CurveFormer++-T, achieves an even higher F-Score (+0.63) and Recall score (+2.1) compared to the single-frame setting.

\textbf{Results on OpenLane Dataset.} For OpenLane dataset, we evaluate CurveFormer++ and CurveFormer++-T on the entire validation set and six different scenario sets.
In Table~\ref{tab:results-openlane-scenario-set}, our CurveFormer++ demonstrates a remarkable improvement compared to dense BEV approaches which have the potential to limit the representation of vertical information. Typically, it outperforms Persformer~\cite{chen2022persformer} by 2.2\% in F-Score across the entire validation set. 

Furthermore, our modification of CurveFormer~\cite{bai2023curveformer} results in a 2. 2\% increase in the F score in the validation set compared to its original version.
In various scenarios, including uphill, downhill, curves, and night conditions, our Transformer-based method provides more accurate results compared to its CNN-based counterpart, Anchor3DLane\cite{huang2023anchor3dlane} which also does not employ a dense BEV module.

Table.~\ref{tab:results-openlane-val-set} shows that CurveFormer++-T achieves more precise predictions on x and z errors, indicating that the incorporation of temporal information can enhance the accuracy of near-region detection results. Additionally, we evaluate PersFormer~\cite{chen2022persformer}, CurveFormer++, and CurveFormer++-T using stability evaluation metrics proposed in~\ref{subsec:metrics} on 200 test video segments. CurveFormer++-T exhibits a lower standard deviation than PersFormer in 96\% of the video segments, and it outperforms CurveFormer++ in terms of standard deviation reduction in 58\% of the video segments. The corresponding results are shown in Fig.~\ref{fig:video_eval_std}.

\subsection{Ablation Study}

In this section, we analyze the effects of the proposed key components via the ablation studies conducted on the subset of OpenLane dataset ($\sim$300 video segments).

\subsubsection{{Single-frame Experiments}}
We conducted validation to assess the effectiveness of CurveFormer++'s components and settings. 

\begingroup
\renewcommand{\arraystretch}{1.5}
\begin{table}[t]
  \begin{center}
  \caption{Ablation of number of anchor point and range prescription module (AncPts: Anchor Points; RR: Range Restriction).}
    \label{tab:ablation:pts}
    \begin{tabular}{ccccccccc}
    \toprule
      \#AncPts & \multicolumn{2}{c}{10} & \multicolumn{2}{c}{40} & \multicolumn{2}{c}{80} & \multicolumn{2}{c}{100} \\
      \midrule
      RR & w/o & w/ & w/o & w/ & w/o & w/ & w/o & w/  \\
      F-Score & 51.8 & 51.9 & 51.3 & \textbf{54.0} & 52.1 & 52.5 & 48.4 & 51.5\\
      \bottomrule
    \end{tabular}
  \end{center}
\end{table}
\endgroup

\begin{table}
    \centering
    \caption{Ablation of batch size and image resolution settings (40 anchor points with range restriction).}
    \begin{tabular}{ccccc}
    \toprule
         Image Resolution&  Encoder&    Params&  FPS&   F-Score\\
         \midrule
        360x480&   \checkmark&   27.36M  &   19.5   &  54.0\\
        360x480&  \XSolid&   23.58M  &   25.1   &53.9\\
        720x960&  \XSolid&    33.16M &   11.4   &\textbf{54.3}\\
    \bottomrule
    \end{tabular}
    \label{tab:ablation of curveformer++ settings}
\end{table}

\textbf{Dynamic Anchor Point Set Design.} In our Curve Transformer Decoder, each 3D lane query is represented as a dynamic anchor point set. The efficiency of dynamic anchor points to extract more accurate features has a crucial impact on the progressive optimization of queries at each decoder layer. To study the effectiveness of our dynamic anchor point set, we conducted experiments that varied the number of anchor points and whether to apply range restriction to the anchor points. The experimental results are listed in Table~\ref{tab:ablation:pts}. 
As anticipated, simply increasing the number of anchor points from 10 to 40 does not improve performance. This is due to the variation in lane lengths, where without range restriction, anchor points may capture irrelevant features. However, when using 40 anchor points with range restriction, we observe a notable 2.7\% improvement in F-Score. Adding more than 40 anchor points does not result in further gains, suggesting that extra anchor points may introduce redundancy, leading to the sampling of duplicated lane features.

\textbf{Network Structures \& Settings.} To simplify the model and reduce computational costs, we replace the original encoder-decoder architecture described in ~\cite{bai2023curveformer} with a decoder-only network. The performances of the two structures are presented in Table~\ref{tab:ablation of curveformer++ settings}. It shows that using the Decoder-only structure has a minimal impact on the performance ($\downarrow$0.1\% F-Score) while reducing computational overhead by 13.8\%. Furthermore, we conducted experiments on CurveFormer++ with different batch sizes and input image resolutions. Among the settings experimented, CurveFormer++ achieves the highest F-Score (54.3\%) and 11.4 FPS on the subset of the OpenLane dataset with 720x960 image resolution. 


\begin{table}[t]
    \centering
    \caption{Ablation of Temporal Propagation Modules (EM:Ego-motion; SA:Self-Attention).}
    \begin{tabular}{ccccc}
    \toprule
        & & Historical Memory&  Operation& F-Score\\
         \midrule
        \ding{202} & - & -& -&51.8 (+0.0)\\ 
        \ding{203} & Fig.\ref{fig:compare_temporal_methods} (a) & Anchor Points&  EM& 52.3 (\hlb{+0.5})\\
        \ding{204} & Fig.\ref{fig:compare_temporal_methods} (b) & Curve Query& SA& 52.7 (\hlb{+0.9})\\
        \ding{205} & Fig.\ref{fig:compare_temporal_methods} (c) & Top-K Curve Query& SA& 53.8 (\hlb{+2.0})\\
        \ding{206} & Fig.\ref{fig:compare_temporal_methods} (d) & \makecell{Top-K Curve Query \\ \& Anchor Points} & \makecell{SA \& EM} & \textbf{54.2} (\hlb{+2.4})\\
    \bottomrule
    \end{tabular}
    \label{tab:ablation of temporal fusion method}
\end{table}

\begin{table}
    \centering
    \caption{Ablation of Temporal Architecture Parameters (720x960 image resolution).}
    \begin{tabular}{ccccc}
    \toprule
         Parameter&  Values&  F-Score\\
                                         
    \midrule                             
                                 
    \multirow{5}{*}{Top-K Num}   & 4& 54.5\\
                                & 6& \textbf{55.4}\\
                                & 8& 53.7\\
                                & 10& 53.9\\
                                & 12& 54.9\\
    \midrule
    \multirow{3}{*}{Historical Frame Length} & 1& 53.4\\
                                        &  2& \textbf{55.4}\\
                                        &  3& 53.5\\
    \bottomrule
    \end{tabular}
    \label{tab:ablation of fusion parameters}
\end{table}

\subsubsection{Multi-frame Experiments}
We now analyze several design choices for CurveFormer++-T.

\textbf{Temporal Propagation Fusion Modules.} In this section, we explore different temporal fusion modules to propagate historical information under the same settings with the single-frame model. $K=6$ is used in the experiments and the results are summarized in Table.~\ref{tab:ablation of temporal fusion method}.

It indicates that utilizing historical anchor points (Fig.\ref{fig:compare_temporal_methods} (a)) and historical curve queries (Fig.\ref{fig:compare_temporal_methods} (b)) produce comparable results and enhance performance by integrating temporal information. Both historical queries and anchor points are effective in improving 3D lane detection outcomes (\ding{203} and \ding{204} v.s. \ding{202}). Moreover, curve queries demonstrate slightly superior results compared to solely employing anchor points (\ding{203} v.s. \ding{204}), implying that latent curve queries may encompass additional information beyond explicit point representation.


By employing historical curve queries and selecting the Top-K curve queries with the highest confidence score (Fig.~\ref{fig:compare_temporal_methods} (c)), an additional enhancement of 1.1\% in F-Score can be achieved (\ding{204} v.s. \ding{205}). This highlights the significance of utilizing informative queries as historical data to encode temporal information. Given that the number of queries often exceeds the actual number of target lanes, some queries may struggle to learn the representation of targets effectively.

Finally, the utilization of Top-K refined sparse curve queries and their corresponding dynamic anchor point sets, Fig.~\ref{fig:compare_temporal_methods} (d)) results in a substantial improvement over the single-frame model, with the F-Score increasing from 51.8\% to 54.2\% (\ding{202} v.s. \ding{206}). The improvement verifies the superiority of our design for the temporal propagation module in capturing and utilizing temporal information effectively.

\textbf{Temporal Fusion Parameters.} Table~\ref{tab:ablation of fusion parameters} presents the results of different temporal fusion module parameters. To meet the real-time requirements of the lane detection task, we limit our experiments to a maximum historical length of 3. 
Using the previous two frames yields better results across different frame lengths. The degradation in performance when using more than two frames may be caused by the presence of a large ego-motion disparity in the time series. 
$K=6$ is the optimal choice for the number of previous frames. This selection aligns with the fact that there are typically six lanes in most real-world scenarios. Maintaining more than six lanes from historical frames may introduce false positives during model propagation. By limiting the number of lanes to six, we ensure more accurate and reliable lane detection results.
The experimental results show that the best performance, with an F-Score of 55.4\%, is achieved by incorporating 6 Top-scoring curve queries, along with their corresponding anchor point sets while considering 2 historical frames.

\section{CONCLUSIONS and LIMITATIONS}
In this paper, we introduce CurveFormer++, a Transformer-based 3D lane detection method. It uses a dynamic anchor point set to construct queries and refines it layer-by-layer in Transformer decoders. 
In addition, to attend to more relevant image features, we present a curve cross-attention module and a context sampling module to compute the key-to-image similarity.  
We also incorporate an anchor point range restriction method to enhance the robustness and adaptability of the model in accurately representing lanes with varying lengths.
Furthermore, our method applies a novel temporal fusion of historical results using sparse curve queries and dynamic anchor point sets. 
In the experiments, we show that CurveFormer++ achieves promising results compared with both CNN-based and Transformer-based approaches. 
While directly using front-view features provides promising results, it also has limitations in scenarios involving long distances, sharp bends, and uneven road surfaces, due to insufficient image features.
Exploring a multi-camera approach offers a promising direction for future research, providing additional context to improve system robustness. Combining multi-modal data, such as detailed 3D LiDAR geometry with perspectives from multiple views, can further enhance the effectiveness of 3D lane detection in challenging scenarios.

\section{ACKNOWLEDEGMENTS}
We would like to thank Zhangjie Fu and Lang Peng for the valuable discussion and comments.

\bibliographystyle{IEEEtran}
\bibliography{refs}

\end{document}